# A Model of Pathways to Artificial Superintelligence Catastrophe for Risk and Decision Analysis

Anthony M. Barrett*,[†] and Seth D. Baum[†]
*Corresponding author (tony@gcrinstitute.org)
[†]Global Catastrophic Risk Institute
http://sethbaum.com * http://tony-barrett.com * http://gcrinstitute.org

**Abstract**

An artificial superintelligence (ASI) is artificial intelligence that is significantly more intelligent than humans in all respects. While ASI does not currently exist, some scholars propose that it could be created sometime in the future, and furthermore that its creation could cause a severe global catastrophe, possibly even resulting in human extinction. Given the high stakes, it is important to analyze ASI risk and factor the risk into decisions related to ASI research and development. This paper presents a graphical model of major pathways to ASI catastrophe, focusing on ASI created via recursive self-improvement. The model uses the established risk and decision analysis modeling paradigms of fault trees and influence diagrams in order to depict combinations of events and conditions that could lead to AI catastrophe, as well as intervention options that could decrease risks. The events and conditions include select aspects of the ASI itself as well as the human process of ASI research, development, and management. Model structure is derived from published literature on ASI risk. The model offers a foundation for rigorous quantitative evaluation and decision making on the long-term risk of ASI catastrophe.

## 1. Introduction

An important issue in the study of future artificial intelligence (AI) technologies is the possibility of AI becoming an *artificial superintelligence* (ASI), meaning an AI significantly more intelligent than humans in all respects. Computers have long outperformed humans in specific domains, such as multiplication. As AI advances, humans are losing in more and more domains, as seen in the recent *Jeopardy!* victory of IBM's Watson computer. These AI can greatly help society, and also pose some risks, but they do not fundamentally change the status quo of humans being in ultimate control of the planet. However, some scholars propose that AIs could be designed that significantly exceed human capabilities, changing the status quo, with results that could be either fantastically good or catastrophically bad for the whole of humanity, depending on how the initial "seed" AI is programmed (Bostrom 2014; Good 1965; Yudkowsky 2008). This proposition is controversial,[1] but if true, the implications are enormous. Given the stakes, the proposition should be taken seriously even if it only has a small chance of being true (Ćirković 2012).

In this paper, we present a graphical model that depicts major pathways to catastrophe caused by ASI seizing control from humanity. We call the model the Artificial Superintelligence Pathways Model or ASI-PATH. The model's pathways involve AIs that recursively self-improve, i.e. AIs that create successively more intelligent AIs via improvements to their hardware or software.[2] The model includes both the human process of building the AI and some aspects of the AI itself.

---
[1] Eden et al. (2012) present a range of perspectives on this controversy and related issues.

The model uses established modeling paradigms from risk and decision analysis, namely fault trees and influence diagrams, in order to depict combinations of events and conditions that could lead to ASI catastrophe, as well as interventions that could avoid ASI catastrophe. Model structures are derived from published literature on ASI catastrophe risk. No attempt is made to quantify model parameters beyond the use of Boolean variables because many of the parameters are highly uncertain; quantifying them rigorously would require significant research beyond the scope of this paper. More generally, given that the broader study of ASI is young and in flux, we expect the ASI-PATH model to evolve in future work. The model presented here can thus be treated as ASI-PATH 1.0.

Even without quantified parameters, the models can serve as communication tools and guides for actions to reduce ASI catastrophe risk. Furthermore, if parameter values are quantified, then the models can be used for quantitative risk analysis, technology forecasting and decision analysis. By characterizing the space of possible AI futures, the models help show, among other things: what could happen; where key points of uncertainty are; which short-term and medium-term precursor events to monitor; and what types of interventions could help avoid harmful outcomes and promote beneficial outcomes. The model presented here thus offers a foundation for rigorous quantitative evaluation and decision making on the risk of ASI catastrophe.

In focusing on ASI catastrophe scenarios, we do not mean to imply that scenarios involving positive ASI outcomes are less important. A global-scale ASI catastrophe would clearly be important, but a global-scale positive ASI event could be similarly important, including because it could help address other global catastrophic risks. That said, much of the modeling presented in this paper can readily be extended to positive ASI scenarios.

The following section provides detailed background on the ASI scenarios, ASI risk-reduction interventions, and the modeling paradigms used in the paper. Section 3 presents the ASI-PATH model by showing a series of interconnected model sections. Section 4 concludes. An Appendix compiles definitions of select technical terms.

## 2. Conceptual Background

### 2.1 AI Catastrophe Scenarios

We focus on scenarios in which AI becomes significantly more intelligent and more capable than humans, resulting in an ASI causing a major global catastrophe. Specifically, we focus on scenarios involving the following steps:

1) Humans create a *seed AI* that is able to undergo recursive self-improvement in either hardware or software.

2) The seed AI undergoes recursive self-improvement, resulting in a *takeoff* of successively more intelligent AIs.

3) The takeoff results in one or more ASIs.

4) The ASI(s) gain *decisive strategic advantage* over humanity, meaning "a level of technological and other advantages sufficient to enable it [the AI] to achieve complete world domination" (Bostrom 2014, p. 78).

---

[2] Seed AI and recursive self-improvement are often used specifically to improvements in software (e.g, Bostrom 2014), but we use the term more generally in deference to the possibility of improvements in hardware (e.g., Sotala & Yampolskiy 2015).

5) The ASI(s) uses their decisive strategic advantage to cause a major global catastrophe, such as (but not necessarily limited to) human extinction.[3]

While other AI scenarios may also be able to bring great benefits or harms to humanity, this type of scenario has been the subject of considerable scholarly and popular attention, enough to merit its own dedicated treatment.

We remain intentionally mute—with one important exception—on what qualifies as a major global catastrophe. The nature of major global catastrophe is a thorny moral issue, touching on issues such as the value of distant future generations.[4] The ASI scenarios raise additional moral issues, such as whether the ASI itself could be inherently morally valuable. How these moral issues are resolved affects how specific seed AIs are to be evaluated, but it does not affect the modeling presented in this paper. The one exception—the one position this paper takes—concerns the mere existence of superintelligence. We assume, for purposes of this paper and without committing our own views, that the mere existence of an ASI with decisive strategic advantage over humanity does not itself qualify as a major global catastrophe. In other words, this paper takes the position that it is not inherently catastrophic for humanity to lose control of its own destiny. We recognize that this is a contestable position, and in using this position in this paper we do not mean to advocate for it. Instead, we use this position in this paper because much discussion of superintelligence effectively takes this position and because it makes for a reasonable starting point for analysis.

Similarly, we have no comment on the relative merits of various concepts for safe (or otherwise desirable) seed AI that have been proposed. These concepts have been discussed using terms such as "Friendliness" (Yudkowsky 2001; 2012), "safe AI scaffolding" (Omohundro 2012; 2014), "AI safety engineering" (Yampolskiy & Fox 2013), and "motivation selection" (Bostrom 2014). Distinguishing between these is also an important task but again beyond the scope of this paper. Instead, we use the generic term "safe AI" throughout the paper without reference to what qualifies as a safe AI and without intending preference for any particular "safe AI" concept. The modeling presented here is of sufficient generality that it could be applied to a wide range of conceptions of desirable outcomes and desirable seed AI.

## 2.2 Risk Reduction Interventions

Models like ASI-PATH can help people make better decisions towards ASI safety—that is, towards reducing the risk of ASI catastrophe. Each model component suggests opportunities to reduce this risk. The following bul    leted list describes the risk reduction intervention options used in ASI-PATH. These are some (but not all) potential risk reduction intervention options, drawing heavily on Sotala and Yampolskiy (2015). The interventions can be made by a range of actors, including governments, companies, universities, funding bodies, civil society, and AI researchers (including both ASI researchers and their colleagues in other branches of AI).

- *ASI research risk review boards*. Review boards would evaluate the riskiness of specific ASI research projects in order to steer research in less risky directions. The review boards could steer research, for example, by disallowing certain projects, directing funds away from riskier projects and towards safer ones, and establishing guidelines for publication. Successful review boards would need to correctly evaluate the riskiness of AI research, which could be difficult (Sotala & Yampolskiy 2015). Review boards might also be limited in the scope of research they review. A board set up by one country may not be able to review research conducted in other countries, though this could be addressed via international treaty (Wilson 2013). Review boards could also overlook ASI projects that operate secretly (Sotala & Yampolskiy 2015).

---

[3] Throughout the paper, we will for the sake of brevity often use "catastrophe" as shorthand for "major global catastrophe".
[4] On the moral importance of global catastrophes, see for example Matheny (2007), Bostrom (2013), Beckstead (2013).

- *Encourage research into ASI safety*. AI safety research can be on any aspect of ASI software or hardware, or on the humans involved in ASI development. ASI safety research could be encouraged, for example, by increasing funding for ASI safety research, creating open source platforms for safe ASI development, or cultivating norms of responsibility among ASI researchers. Encouraging research can be an easier intervention to perform than something like review boards, because encouraging research can be done informally and outside institutional procedures. However, informal encouragement can sometimes be less effective when it lacks institutional clout. As with research risk review boards, encouraging research requires knowing whether a particular activity would increase or decrease ASI risks.

- *Enhance human capabilities*. Human enhancement in this context refers to anything that could help humans keep up with AI, such that the AI does not gain decisive strategic advantage even during or after takeoff. Human capabilities could be enhanced, for example, via brain-computer interfaces or mind uploading. Human enhancement could reduce some ASI risks, but it could increase other ASI risks (e.g., if enhanced humans create unsafe ASI) or cause other problems (e.g., moral problems associated with modifying human nature). Additionally, human enhancement could require advanced technology that may not be developed before the seed AI, in which case human enhancement would not help with ASI safety.

- *AI confinement*. AI confinement measures are restrictions built into the AI's hardware or software that limit the AI's ability to affect the rest of the world so that it does not gain decisive strategic advantage.[5] Confinement can be applied to the seed AI, the recursively self-improving AIs, or the ensuing ASI(s). For example, AI software could be designed such that the AI can only answer questions; this is known as "Oracle AI" (Armstrong, Sandberg & Bostrom 2012). Another type of confinement would keep AI completely disconnected from the world, for example to test the AI's safety (Yampolskiy & Fox 2013).[6]

- *AI enforcement*. AI enforcement occurs when one or more AIs prevent other AI(s) from gaining decisive strategic advantage. As with confinement, AI enforcement can be applied to seed AI, self-improving AIs, or ASIs; the enforcement can also be performed by any of these AIs, though more intelligent AIs may tend to be more effective enforcers. For AI enforcement to work, the AIs with strategic advantage must be motivated and able to stop other AIs. For scenarios with multiple AIs, there could be a first-mover advantage, such that the first AI(s) gain and keep strategic advantage. In such scenarios, the first AI(s) must be designed to conduct enforcement.

---

[5] Confinement and AI enforcement are types of containment, with containment being any measure to restrict the AI's ability to affect the rest of the world such that the AI does not gain decisive strategic advantage. Other types of containment are measures taken by the rest of society and not built into the AI itself. We do not model these other types of containment as risk reduction interventions.

[6] Goertzel and Pitt (2012) argue that, under some circumstances, connecting the AI to the world could increase safety. Even in such circumstances, keeping the AI disconnected would still qualify as confinement: confinement does not necessarily increase safety.

## 2.3 Fault Tree and Influence Diagram Modeling

ASI-PATH uses conventions from two established modeling paradigms: fault trees and influence diagrams. Fault trees offer a simple but powerful method for modeling sets of interrelated possible scenarios. Fault trees have been used in modeling risks in a wide variety of contexts, including nuclear power plant safety (Rasmussen 1981), homeland security risk (Cheesebrough 2010), nuclear war (Barrett, Baum & Hostetler 2013; Davis 1989), and other kinds of systems (Bedford & Cooke 2001; Kumamoto & Henley 1996).

Fault trees represent the ways that events and conditions could combine to be at fault for causing a particular outcome, in this case an ASI catastrophe. Each node in the tree represents a particular event, such as an attempted creation of a seed AI, or a condition, such as the presence of some technology keeping the AI safe. The "top event" node in the tree represents the scenario outcome. Below the top node, the tree branches out with additional nodes corresponding to different ways that the top node can be reached. Each layer in the tree represents the combination of events and conditions that could lead to the outcome in the layer directly above it. Nodes are connected by Boolean logic gates, such as OR, AND, and NOT gates. The tree thus shows a set of possible scenarios that could result in the top event.

Influence diagrams represent the ways that certain events can influence subsequent events. The events can be random variables, determined by factors external to the model, or decision variables, determined by decisions informed by the model. Influence diagrams thus can be thought of as Bayesian networks with the addition of decision nodes that represent variables whose values can be chosen instead of having a random value (Pearl 1988). Influence diagrams have been used for decades in decision analysis to inform business, government and other policy decisions under conditions of uncertainty (Clemen & Reilly 2001; Edwards, Miles & von Winterfeldt 2007). ASI-PATH uses influence diagrams to model the risk reduction interventions described in Section 2.2.

The figures in this paper use standard visual conventions for fault trees and influence diagrams. Connector lines are used to connect nodes representing events and conditions. Vertical connectors with no arrows are used for fault trees; the lower node is at fault for causing the upper node. Horizontal connectors with arrows are used for influence diagrams; the left node influences the right node. Intervention options appear as unrounded rectangles with white text on black background; in influence diagram terminology, these are called decision nodes. Other (non-intervention) events appear as rounded rectangles with black text on white background; these are called random-variable nodes. Some figures use both fault tree and influence diagram conventions, as well as both decision nodes and random-variable nodes.[7]

---

[7] The use of white text on black background is not a standard influence diagram visual convention. We use it to further distinguish decision nodes from random-variable nodes in our figures.

## 3. The Artificial Superintelligence Pathways Model

### 3.1 Model Overview

Figure 1 presents the first three layers of ASI-PATH. The first two layers of Figure 1 should be fairly intuitive: an ASI catastrophe occurs if a seed AI takes off, resulting in a superintelligence with decisive strategic advantage over humanity, AND if the ASI's actions are unsafe, such that it uses the decisive strategic advantage to cause catastrophe.[8]

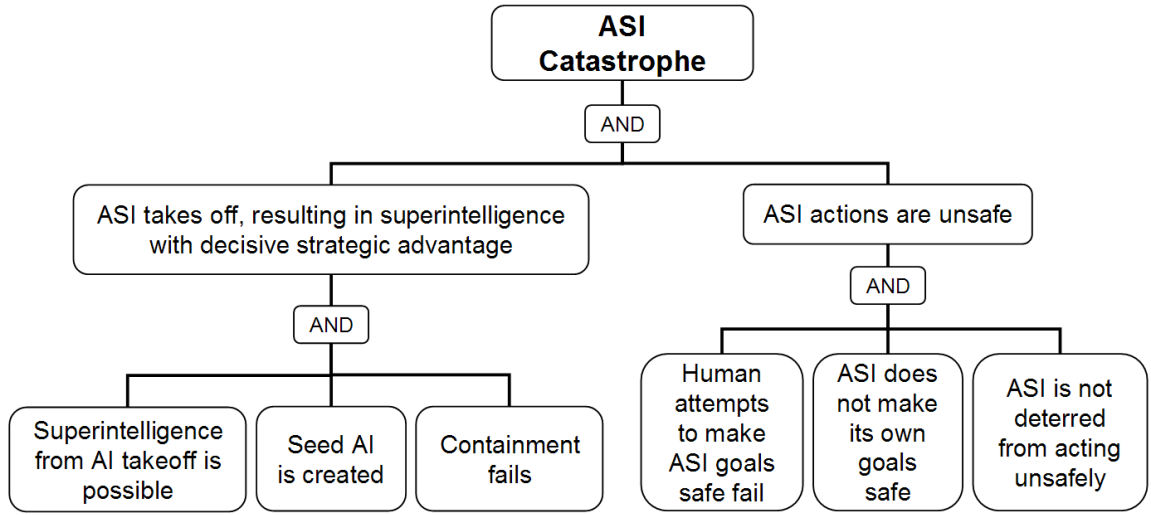

**Figure 1: ASI-PATH, fault tree top three layers**

The third layer in Figure 1 contains more model detail: six conditions that must all be in place in order for ASI catastrophe to occur. If any of these conditions are missing, then there will be no AI takeoff, or the resulting ASI will not cause catastrophe. Starting on the left side: In order for a seed AI to be created and have the capability to gain decisive advantage, three conditions must all be met:

*Superintelligence from AI takeoff is possible.* There is some debate about whether it could ever be physically possible for ASI to be created via recursive self-improvement. There is also debate on whether humans are capable of creating such ASI. For example, Bringsjord et al. (2012 pp. 403-406) argue that it is far from certain that humans would ever be able to create AI with human-level intelligence, or that such an AI would be able to recursively create more-intelligent AI to arrive at superintelligence.

*Seed AI is created.* For ASI catastrophe to occur, there must be a project in which humans build a seed AI capable of becoming superintelligent via recursive self-improvement. Such an AI is presumably not going to spontaneously materialize out of thin air. There are a variety of ways in which ASI projects could build a seed AI (Section 3.2).

*Containment fails.* If containment fails, then the ASI gains decisive strategic advantage. Containment measures could include takeoff limits, such that the seed AI does not begin recursive self-improvement, or such that successive AIs do not continue recursive self-improvement so as to become superintelligent. Other containment measures could prevent a post-takeoff ASI from gaining decisive strategic advantage, for example via confinement, human enhancement, or AI enforcement (Section 2.2).

---

[8] The model can be adapted for analyzing safe ASI by switching the layer 2 right node to "ASI actions are safe"; much of the lower-layer modeling (discussed below) will be similar.

Moving to the right side of Figure 1, third layer: In order for an ASI to have unsafe goals—in order for it to not use its decisive strategic advantage to cause catastrophe—three conditions must all be met:

*Human attempts to make AI goals safe fail.* Goal safety can potentially be built into an ASI by the AI designers. Indeed, some people are currently calling for ASI developers to make their AIs' goals safe (Bostrom 2014; Yudkowsky 2012), and there is ongoing research into various ways to make ASI goals safe (Goertzel & Pitt 2012; Yampolskiy & Fox 2013). There is some debate about how to design safe ASI goals—more on this below.

*AI does not make its own goals safe.* Several people have suggested that AIs could make their own goals safe even if the goals were not designed to be safe (Hall 2011; Waser 2008). One possible reason could be that "minds which are intelligent enough will, due to game-theoretical and other considerations, become altruistic and cooperative" (Sotala & Yampolskiy 2015 p. 21). Others have disputed this point, arguing that without significant dedicated effort, catastrophe is likely to result (e.g., Bostrom 2014; Omohundro 2008; Omohundro 2012; 2014; Yampolskiy & Fox 2013; Yudkowsky 2001; 2012). For example, Omohundro (2008) argues that almost any simple goal, such as playing chess well, could be unsafe. The AI would cause catastrophe in its relentless pursuit of that goal, such as by hoarding the world's resources to play a better game of chess.

*ASI is not deterred from acting unsafely.* To deter is to persuade another agent to refrain from taking some action by convincing the agent that the action is not in their own interest. Even if an ASI has decisive strategic advantage and unsafe goals, it potentially could still be deterred from using its advantage to cause catastrophe. However, Bostrom (2014) argues that an ASI with decisive strategic advantage would be undeterrable.

We now fill in some additional model details with additional fault tree layers, first for the possibility of limits to AI takeoff, then for human efforts to make the AI safe. The subsequent figures show some major relationships between decision nodes (rounded rectangles, black text on white background) and random-variable nodes (unrounded rectangles, white text on black background). The selection of relationships presented is not intended to be comprehensive. The decision nodes can influence other random-variable nodes besides those shown in the figures. There can also be other decision nodes representing other interventions not modeled in this paper. What is shown is intended only to be illustrative of how decision nodes can be used to incorporate risk reduction interventions in the ASI-PATH model.

### 3.2 Seed AI Creation

A critical question is how and when an AI research and development project could build a seed AI. When the seed AI is created matters for several reasons. Earlier creation could make takeoff slower and easier to contain (Section 3.3), but earlier creation also gives less opportunity to design a safe seed AI.

Figure 2 presents additional detail on seed AI creation, extending the "Seed AI is created" node from Figure 1. Figure 2 shows two major types of seed AI: AI of novel design and whole brain emulation. Figure 2 also has a catch-all "other seed AI" branch for any types of seed AI that do not fit into the two major types. This section of the model draws heavily on Bostrom (2014, ch. 2).

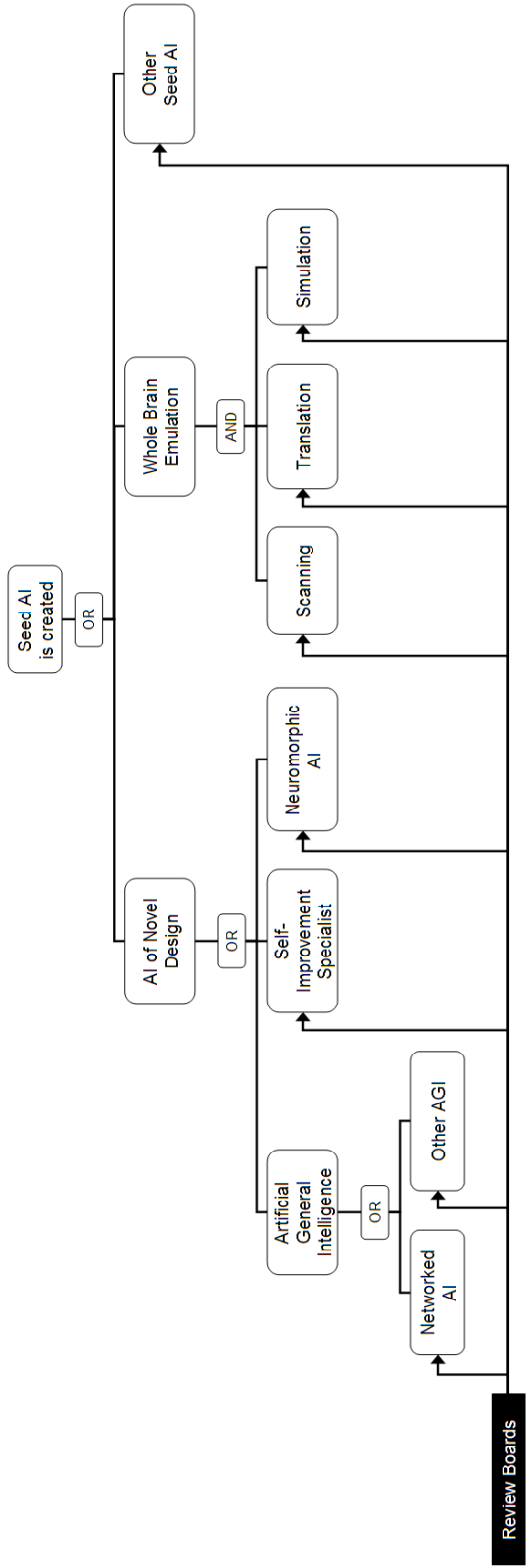

**Figure 2: Detail on seed AI creation**

*AI of novel design* refers to seed AI whose design is at least novel relative to the human brain. In other words, the seed AI design is at least in part not based on the human brain. Four different pathways to seed AI of novel design are modeled:

*(1) Artificial general intelligence (AGI)* in this context is seed AI with approximately human-level or greater general intelligence—that is, at approximately human-level or greater intelligence in all respects (e.g., Goertzel & Pennachin 2007).[9] Since humans are capable of programming better AIs, an AI at least as smart as humans in all respects would also be able to program better AIs, thereby beginning a recursive self-improvement process (Bostrom 2014; Sotala 2012). The AGI model branch contains a lower layer to model one pathway to AGI that has been identified, which comes from networks or organizations of AIs, for example if "the Internet might one day 'wake up'" (Bostrom 2014, p. 49). The lower layer also includes a catch-all "other AGI" node.

*(2) Self-improvement specialist* is seed AI that is "sub-human in most respects, but with a domain-specific talent for coding and AI research" (Bostrom 2014, p. 29). In other words, the AI is a specialist in recursive self-improvement. For example, Sotala (2012) and Bostrom (2014) discuss ways in which superintelligence could be achieved by AI with initially subhuman general intelligence but greater than human ability to take control of other networked systems, make copies of itself, and to coordinate between copies of itself.

*(3) Neuromorphic AI* is seed AI that is based in part, but not entirely, on the human brain: a partial brain emulation. For example, neuromorphic AI could be based on human patterns of cognition while lacking human motivations; the difference in motivations could make it harder to achieve goal safety.

*(4) Other novel design*, a catch-all node for any pathways to novel seed AI designs that do not fit into the three identified pathways.

Moving on to the second major branch of Figure 2—the second major type of seed AI:

*Whole brain emulation* is seed AI that emulates the entirety of the human brain (Alstott 2014; Sandberg & Bostrom 2008). Because the human brain is capable of designing AIs, a whole brain emulation AI presumably could also design AIs. A whole brain emulation AI could also self-improve by using more and faster hardware, something that biological human brains cannot readily do. By imitating the human brain, whole brain emulation may require less intellectual breakthrough than seed AI of novel design—it "relies less on theoretical insight and more on technological capability" (Bostrom 2014, p. 33).

Whole brain emulation requires each of three steps to occur:

*(1) Scanning*, in sufficiently detail, of one or more whole human brains. The brain scan creates an image of the whole brain to be emulated.

*(2) Translation*, using data analysis and image processing, to construct digital representation(s) of the brains neuronal network and/or any other relevant information. The translation thus converts the scanned image into a format that can be emulated.

---

[9] The term AGI can refer to more than just seed AI. Indeed, an ASI built from a seed AGI would likely also be an AGI, as the AI is unlikely to lose general intelligence as it undergoes recursive self-improvement.

*(3) Simulation*, which creates a whole brain emulation seed AI based on the whole brain translation.

Figure 2 shows the decision node "Review Boards" (i.e., ASI research risk review boards) influencing each of the model section bottom nodes. Review boards could evaluate the safety of each type of seed AI development project. For projects found to be unsafe, review boards could recommend that the projects stop their work or that funders stop funding the projects. Such a board would be analogous to, for example, the review of risky biomedical research ("dual use research of concern" or DURC) by the United States Department of Health and Human Services (HHS; see e.g., OBA 2012; Patterson, Tabak, Fauci, Collins & Howard 2013). The DURC review helps determine whether HHS (which includes the National Institutes of Health) funds the research and what safety procedures the research must follow.

Figure 2 shows separate arrows pointing to each of the bottom nodes instead of one arrow pointing to the top node ("Seed AI is created") because the influence of Review Boards can be different for each bottom node. That is, Review Boards could have different influence on different types of seed AI project. For example, perhaps Review Boards are more effective for neuromorphic AI, which could require large dedicated projects, and less effective for networked AI, which could emerge from entrenched infrastructure (e.g., the Internet). The use of separate arrows illustrates how this model section would be coded, with different influence relations for different bottom nodes.

## 3.2 AI Containment

Containment is defined in this paper as any measure that restricts an AI such that it does not result in an ASI gaining decisive strategic advantage. The restriction can be on seed AI, self-improving AI, or ASI. Figure 3 models some key details about containment, extending the "Containment fails" node in Figure 1.

Working down from the top of Figure 3, containment fails if takeoff limits fail or if superintelligence containment fails. In other words, an ASI gains decisive strategic advantage if a seed AI takes off, becoming superintelligent, and if the resulting ASI is not contained.

Starting on the left branch of Figure 3, AI takeoff limits fail if either hard takeoff limits fail or soft takeoff limits fail, simply because these are the two types of takeoff:

*(1) Hard takeoff limits fail*. A hard takeoff proceeds rapidly, with no opportunity for human intervention. Figure 3 models three hard takeoff scenarios (i.e., failures of hard takeoff limits): (1) hardware quantity, in which the AI takes off rapidly due to an abundance of available hardware (also known as hardware overhang; see Sotala and Yampolskiy 2015); (2) hardware quality, in which the AI rapidly develops faster hardware (also known as speed explosion; see Sotala and Yampolskiy 2015); and (3) software quality, in which the AI rapidly develops smarter AI software (also known as intelligence explosion; see Sotala and Yampolskiy 2015).

*(2) Soft takeoff limits fail*. A soft takeoff proceeds slowly, permitting significant human intervention in the takeoff process. Figure 3 models three soft takeoff scenarios (i.e., failures of soft takeoff limits): (1) hardware quantity, in which the AI takes off due to gradually increasing its amount of available hardware;[10] (2) hardware quality, in which AI takes off due to gradually develops faster hardware; and (3) software quality, in which the AI gradually develops smarter AI software.

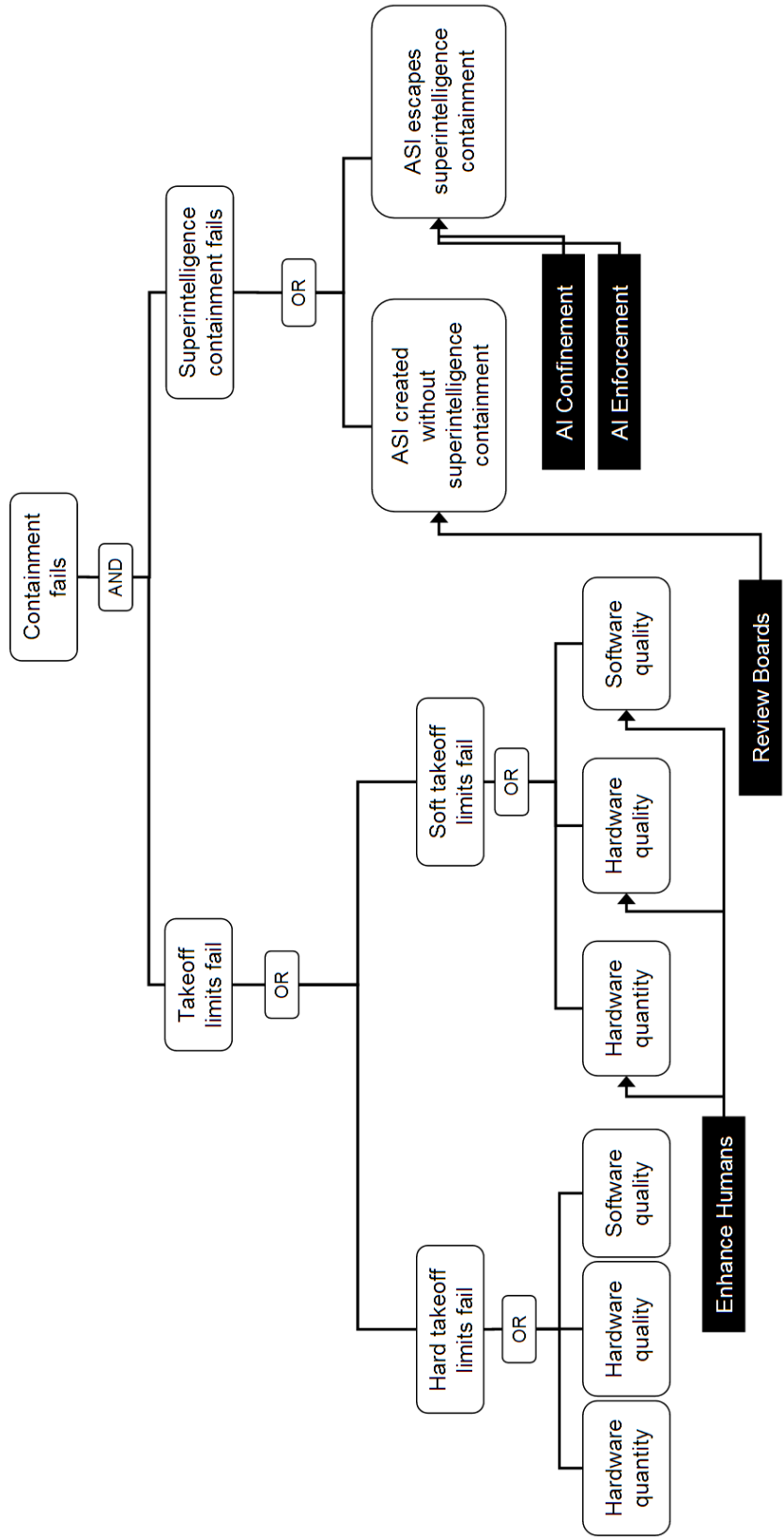

**Figure 3: Detail on containment failure**

Takeoff speed is important because it strongly affects how much opportunity there will be to manage the recursive self-improvement process. If a hard takeoff is more likely, then it is especially important to design the seed AI so as to achieve desirable outcomes, because there will be no opportunity to adjust the AI once takeoff begins. Shulman and Sandberg (2010) and Goertzel and Pitt (2012) suggest that a soft takeoff would be more likely if the seed AI is created earlier, because the hardware, programming languages, etc. available for the AI would tend to result in slower seed AI. Thus to the extent that the seed AI creation model section (Section 3.2) can indicate when seed AI may be created, this can inform the success of containment.

Figure 3 shows the decision node "Enhance Humans" (i.e., enhance human capabilities) influencing the three random-variable nodes below "Soft takeoff limits fail". In principle, enhancing human capabilities could help with a wide range of AI safety measures, but it could be especially useful for limiting soft takeoff. This is because soft takeoff involves AIs gradually gaining capabilities; enhanced humans may be able to keep up with the AI during much of the takeoff, ensuring that no unsafe ASI gains decisive strategic advantage. In comparison, a hard takeoff would by definition occur too quickly for even enhanced humans to keep up.

As with Figure 2, Figure 3 shows separate arrows pointing to each of the bottom nodes instead of one arrow pointing to the top node ("Soft takeoff limits fail") because the influence of Enhance Humans can be different for each bottom node—for each type of soft takeoff. For example, perhaps Enhance Humans is more effective for containing hardware quality soft takeoff than for containing software quality soft takeoff, because enhanced humans could control the AI's access to the physical equipment necessary for hardware production. As with Figure 2, the use of separate arrows illustrates how this model section would be coded, with different influence relations for different bottom nodes.

Moving on to the right branch of Figure 3, superintelligence containment fails if there is no superintelligence containment or if the AI escapes whatever superintelligence containment is used:

*(1) ASI is created without superintelligence containment.* Some ASI development projects might not use any superintelligence containment. This could happen for example if the developers have no ability to use superintelligence containment, or if they believe the superintelligence containment would not work, or if they want the ASI to gain decisive strategic advantage. Whatever the reason, ASI cannot be confined if there is no superintelligence containment.

Figure 3 shows the decision node "Review Boards" (i.e., ASI research risk review boards) influencing the node "ASI is created without superintelligence containment". Review boards could monitor ASI R&D projects and require them to have some sort of containment.

*(2) ASI escapes superintelligence containment.* Superintelligence containment might not be designed and implemented well, and even world-class superintelligence containment may fail to contain some ASIs. Bostrom (2014 pp. 129-131, 145-148, 283) discusses many potential ways an ASI could escape containment, including persuading or tricking any gatekeepers who can remove containment, using methods of interacting with the outside world that containment designers failed to anticipate, or pretending to malfunction so that the containment is at least temporarily disabled.

---

[10] One might argue that a "hardware quantity" soft takeoff is unlikely on grounds that if an AI is able to access more hardware than its designers intended, it would likely be able to access much or all of the entire world's hardware all at once, causing a hard takeoff. While quantifying scenario probabilities is beyond the scope of this paper, such probabilities can be plugged directly into the model.

Figure 3 shows the decision nodes "AI confinement" and "AI enforcement" influencing the random-variable node "AI escapes superintelligence containment". AI confinement and AI enforcement are two types of superintelligence containment, so this influence relation follows from basic logic.

### 3.3 Human Attempts at Making ASI Goals Safe

Modeling ASI goal safety is a subtle and challenging task because there is no clear consensus on what it takes to make ASI goals safe. The study of ASI goal safety is, we think it is fair to say, in early stages, with a lot of uncertainty and disagreement. Because of this, one cannot model ASI goal safety with much specificity, or at least not with much confidence about the specifics. Therefore, the models in this section are intended as exploratory, in order to illustrate some of what is known and to promote open discussion with an eye toward future work. We expect to change these models pending such discussion and pending results of ongoing research on ASI goal safety.

#### 3.3.1 General Failure Types

Figures 4-7 model general types of failure of human attempts at making ASI goals safe. Figure 4 models two major lines of current thinking about how humans could make ASI goals safe. The two lines of thinking are the left and center nodes of layer 2, with the right node being a catch-all category for any other lines of thinking:

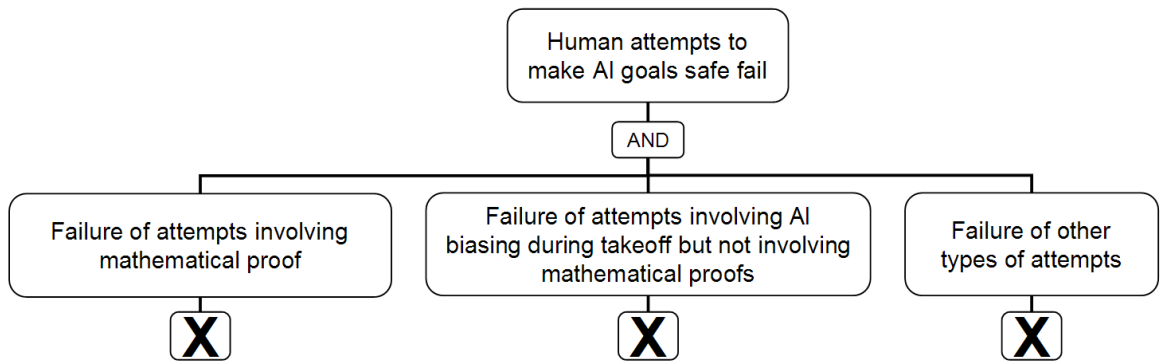

*Figure 4: Two lines of thinking on human attempts at making AI goals safe*

*(1) Failure of attempts involving mathematical proof.* Some people involved in ASI goal safety research have called for goal safety measures to be built into the seed AI, and furthermore for the correctness of these measures to be mathematically proven before the AI is launched. For example, ASI developers could build in *goal stability*, such that the AIs goals would not change as it undergoes recursive self-improvement. It may be possible to mathematically prove goal stability prior to launch.[11] Mathematical proofs may be able to give AI designers and other observers a high confidence that the ASI's goals would be safe, though it would not guarantee safety with 100% probability (Brundage 2014; Muehlhauser 2013). Requiring mathematical proofs before launch could also delay launch, by creating an additional hurdle that must be cleared before the seed AI is launched. Given that the failure of ASI goal safety could result in major global catastrophe, requiring mathematical proofs before launching is thus in the spirit of the precautionary principle, a major policy paradigm for risky emerging technologies.[12] Finally, note that this line of thinking does not exclude goal safety measures that occur after seed AI launch.

---

[11] For discussion of this and related ideas, see for example Yampolskiy and Fox (2013), Omohundro (2012; 2014), Soares and Fallenstein (2014), and Russell et al. (2015). Details of failure modes for these ASI goal safety measures could be modeled in extensions of Figure 3.

*(2) Failure of attempts involving AI biasing during takeoff but not involving mathematical proofs*. Other people in ASI goal safety research have argued against mathematical proofs prior to seed AI launch, and instead argue for biasing the AI's goals towards safety as the AI recursively self-improves. This line of thinking argues that the AI's goals could be biased through human interaction with the AI as it grows and learns, using reward and punishment from a trainer/biaser in a fashion somewhat analogous to what human children experience from their parents (Goertzel & Pitt 2012). Such AI biasing requires a soft takeoff; otherwise human interaction during takeoff would be impossible. Because mathematical proofs before launch are not sought, this line of thinking could enable earlier launch of the seed AI. There could be reasons for favoring earlier launch, for example in the context of a race between competing AI projects. However, supporters of requiring mathematical proofs have argued that the belief that proofs are unnecessary might give AI designers a dangerous misperception that an AI's goals would remain safe outside its training environment (Sotala & Yampolskiy 2015 pp. 21-23).

*(3) Failure of other types of attempts*. This node is a catch-all for any attempt at ASI goal safety that does not involve either mathematical proofs or biasing during takeoff. While other types of attempts are, to our knowledge, not prominent in ASI goal safety literature, there may nonetheless be other types of attempts that could achieve ASI goal safety.

Each of the three nodes in the second layer of Figure 4 have a lower box labeled X. The X box is a model module, essentially an abbreviation for a repeating model section. The abbreviation X is used to conserve space in the diagram, with no implications for the underlying model logic. We refer to this as a model module by analogy to modular computer programming. If the model is written in computer code, then modular programming could be used for X. The details of X are shown in Figure 5.

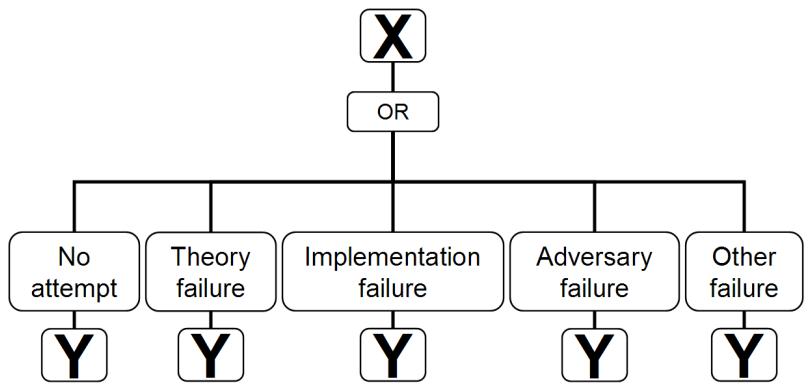

**Figure 5: Explanation of the X module used in Figure 4**

---

[12] On the precautionary principle in the context of catastrophic risk, see for example Posner (2004); Sunstein (2009).

Figure 5 shows that the X module has five nodes connected by an OR gate, each with a new module Y beneath it. The five nodes are:

*No attempt*, meaning no attempt is made at ASI goal safety using the types of attempt shown in Figure 4.

*Theory failure*, meaning a failure to achieve ASI goal safety due to some flaw in the theory of the type of goal safety attempted. Theory failure could be due, for example, to a philosophical mistake, such as selecting a goal that turns out to be unsafe, as in an ASI that causes catastrophe in the relentless pursuit of better chess (Omohundro 2008).

*Implementation failure*, meaning a failure to correctly implement the ASI goal safety technique in the actual AI. Implementation failure could be due to such things as a coding error or a hardware glitch.

*Adversary failure*, meaning the ASI goals become unsafe due to interference by some sort of adversary. The adversary could be a rogue member of the AI R&D team, an external party such as a competing ASI team or a misguided governing body, or another AI, as in AI enforcement.

*Other failure*, a catch-all for failure modes not covered in the other four nodes.

Each of the five Figure 5 nodes has a module Y beneath it. Module Y is shown in Figure 6. Module Y contains two nodes connected by an OR gate: failure before and during takeoff. Some ASI goal safety measures must be taken before takeoff, such as those involved in designing and coding the seed AI. Other safety measures must be taken after takeoff, such as those involving interacting with the AI so that its self-improvement proceeds in a safe direction.

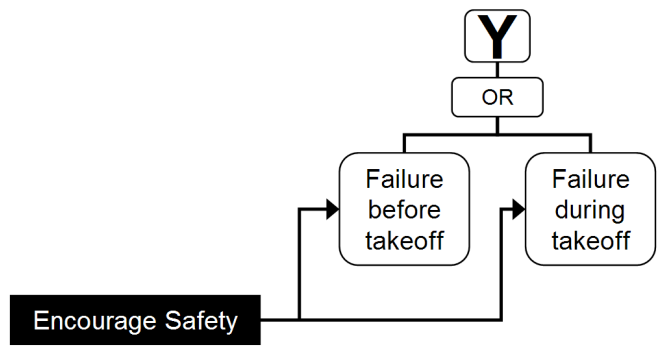

Figure 6: Explanation of the Y module used in Figure 4

Figure 6 shows the decision node "Encourage Safety" (i.e., encourage research into AI safety) influencing the bottom random-variable nodes in Figure 6. These influence relations indicate that Encourage Safety could influence all types of human attempts at making ASI goals safe. ASI goal safety is not the only type of AI safety that one could encourage, but it is a notable type. Active efforts exist to promote ASI goal safety research, accompanied by arguments that this is an especially important task for improving ASI outcomes (e.g., Muehlhauser & Bostrom 2014).

The use of separate arrows pointing to each of the bottom nodes in Figure 6 is analogous to the use of separate arrows in Figures 2 and 3: to indicate that the influence can be different for different bottom nodes. Because of the X and Y modules, the bottom nodes in Figure 6 are each repeated 15 times, meaning that there are 30 different Encourage Safety influence relations. Code of this model section could have different relations for each of the 30 bottom nodes. For example, perhaps it is relatively effective to encourage safety for "Failure of attempts involving mathematical proof"—"No attempt"—"Failure before takeoff" because before takeoff, it is relatively easy to encourage ASI developers to make at least some attempt at mathematical proofs, and perhaps it is relatively ineffective to encourage safety for "Failure of attempts involving AI biasing during takeoff but not involving mathematical proofs"—"Theory failure"—"Failure during takeoff" because after takeoff, there might not be enough time to develop adequate theory.

### 3.3.2 Specific Safety Measures

Figure 7 presents a different format for modeling human attempts to make ASI goals safe. Instead of organizing the attempts by lines of thinking and general failure types, as in Figures 4-6, Figure 7 organizes the attempts by the specific safety measures the attempts could involve. Figure 7 thus aims to model the details of what goal safety requires. The goal safety measures are grouped into two categories: measures built into the seed AI before its launch and measures taken during takeoff. This categorization assumes that no human attempts could succeed after takeoff, since then the AI would be superintelligent and presumably impervious to human interventions.

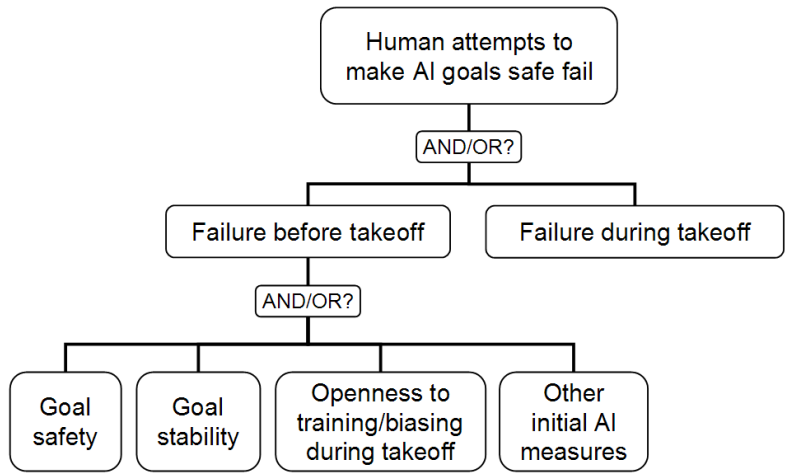

**Figure 7: AI goal safety measures**

Figure 7 includes four types of safety measures that could be built into the seed AI before takeoff: (1) goal safety, meaning safety measures that result in the post-takeoff superintelligence having safe goals;[13] (2) goal stability, meaning the AI keeps the same goals throughout recursive self-improvement; (3) openness to training/bias during takeoff, such as to permit "attempts involving AI biasing during takeoff but not involving mathematical proofs" (Figure 3 and above);[14] and (4) other seed AI measures, a catch-all category for any other safety measures that could be built into the seed AI.

Figure 7 uses two "AND/OR?" logic gates. The reason for this unusual logic structure is because it is not clear whether the logic gates should be ANDs or ORs, and indeed they potentially could be either. The "AND/OR?" gates are thus a representation of model uncertainty, analogous to the use of probability distributions for parameter uncertainty. Likewise, the gates can be coded by considering each combination of AND and OR, perhaps with a probability assigned to each combination.

---

[13] Goal safety does not require the seed AI or subsequent pre-superintelligence AIs to have safe goals—thus included here are ideas like "coherent extrapolated volition" (Yudkowsky 2004) in which AI acquires safe goals during takeoff.

[14] See also Soares (2015) for a characterization of openness to training: "corrigibility", which an AI system would possess "if it cooperates with what its creators regard as a corrective intervention, despite default incentives for rational agents to resist attempts to shut them down or modify their preferences."

For the top logic gate:

- An AND gate would be used if, in order for the top node failure to occur, it is necessary for both layer 2 node failures to occur. That would mean that ASI goal safety could be achieved either via building measures into the seed AI or by taking measures during takeoff. For example, if the seed AI can be made safe before launch, then additional measures during takeoff might not be necessary. Indeed, if the takeoff is hard, then there may not be opportunity for additional measures during takeoff, and goal safety would depend on safety measures built into the seed AI.

- An OR gate would be used if, in order for the top node failure to occur, it is sufficient for either of the layer 2 node failures to occur. That would mean that AI goal safety would require measures built into the seed AI and measures taken during takeoff. For example, for an AI to be biased towards safe goals during takeoff, it might need certain safety measures built into the seed AI, such as openness to training/biasing during takeoff.

For the bottom logic gate:

- An AND gate would be used if, in order for the top node failure to occur, it is necessary for all four bottom node failures to occur. That would mean that pre-takeoff ASI goal safety could be achieved via any one of the four safety measures. For example, if the seed AI is open to bias/training during takeoff, then it may not need to have safe or stable goals before takeoff, as its goals can be adjusted during takeoff.

- An OR gate would be used if, in order for the top node failure to occur, it is sufficient for any of the four bottom node failures to occur. That would mean that AI goal safety would require all four safety measures to be built into the seed AI before takeoff. For example, goal safety may be useless if the goals are not also stable, because the goals would become unsafe during takeoff.

## 4. Conclusion

It is far from certain that ASI will ever be built, or that, if it is built, it will result in catastrophe. But given the stakes—given the extreme severity of the catastrophe that could result—ASI catastrophe risk merits careful analysis and management. The ASI-PATH model presented in this paper is one step towards ASI risk analysis and risk management decision-making. ASI-PATH uses fault trees and influence diagrams, which are established risk and decision analysis methodologies with a long history of success at informing risk management decisions. This history, and the experience it has generated, is leveraged in ASI-PATH for the analysis of ASI risk.

One might think that ASI is such a novel risk that established methodologies would not be suitable. ASI risk does indeed have some significant novel characteristics, in particular the possibility of humans losing control to the ASI. However, core aspects of ASI risk are broadly similar to other risks. ASI risk involves sets of pathways, events, and conditions involving interactions between humans and the technologies they develop. Analysis of ASI catastrophe pathways is thus not fundamentally different from analysis of pathways to other technology catastrophes. This is good news, because it makes analysis of ASI catastrophe risk an easier endeavor.

That said, ASI catastrophe risk remains poorly understood. These are early days for the study of ASI risk. The ASI-PATH model presented here and the understanding of ASI risk it represents are both likely to evolve as the field matures. This is to say that the ASI-PATH model is wrong —but then again, all models are wrong. Models like ASI-PATH can still be useful by helping people better understand the underlying risk, including the major points of uncertainty. For example, the modeling in this paper suggests that one key uncertainty is the combination(s) of measures needed to make ASI goals safe, hence the model uncertainty ("AND/OR?" gates) depicted in Figure 7. To the extent that future research could reduce the uncertainty, it would be of value for understanding ASI risk and informing ASI risk management decisions.

An important task for future work is the quantification of ASI-PATH parameters using real-number values instead of Boolean logic. Doing so would enable the evaluation of ASI risk and specific risk reduction interventions, towards the goal of taking the most effective actions to reduce ASI risk. Quantifying ASI-PATH parameters is a delicate task due to the significant uncertainty that exists about parameter values, but this only means that quantification should proceed carefully, not that it should not proceed. Indeed, given the stakes associated with ASI catastrophe and the corresponding opportunities for reducing ASI catastrophe risk, the entire study of ASI catastrophe risk should be a significant research priority.

**Appendix A: Glossary of Key Terms**

This paper uses a lot of terminology, including many ordinary words used to refer to specific technical concepts. To aid the reader, here is a glossary of key terms as they are used in this paper. We have strived for consistency with prior literature, though this paper may have some unique terminology or definitions.

An *AI system* includes the AI software ("the AI itself") and the hardware it runs on, including hardware it uses for sensory input (e.g., cameras), external communications (e.g., monitors displaying text messages), and any other external interactions (e.g., robot arms).

*Confinement* is any measure built into an AI system that restricts its ability to affect the rest of the world such that the AI does not gain decisive strategic advantage. Confinement measures by definition do not affect the rest of the world, or at least do not affect it to any significant extent. For example, putting the AI in a metal box to shield its radio emissions could be considered confinement even if the rest of the world would be affected by having to purchase and install the box. Confinement measures can be either hardware (e.g., a metal box) or software (e.g., code that prevents the AI from physically moving objects).

*Containment* is any measure that restricts an AI system's ability to affect the rest of the world such that the AI does not gain decisive strategic advantage. Containment includes measures built into the AI system (i.e., *confinement*) as well as measures involving the rest of the world (e.g., AI *enforcement*).

*Goal safety* refers to safety measures built into AI software giving the AI goals such it does not use decisive strategic advantage to cause major global catastrophe. AI goal safety is similar, but not necessarily equivalent, to established concepts such as "Friendliness" (Yudkowsky 2001; 2012) and "motivation selection" (Bostrom 2014).

*Safety* is any measure applied in any way to an AI system that reduces the risk of the AI causing a major global catastrophe. Safety thus includes measures to prevent humans from launching a seed AI (e.g., research review boards), measures to prevent the seed AI or any subsequent AI from gaining decisive strategic advantage (e.g., *takeoff limits*), and measures to prevent a superintelligent AI from using decisive strategic advantage to cause major global catastrophe (i.e., *goal safety*).

*Takeoff limits* are measures that restrict an AI's ability to recursively self-improve. Takeoff limits can be used for *containment* or for other purposes (e.g., the scientific study of takeoff dynamics).

**Acknowledgments**

Thanks to Daniel Dewey, Nate Soares, Luke Muehlhauser, Miles Brundage, Kaj Sotala, Roman Yampolskiy, Eliezer Yudkowsky, Carl Shulman, Jeff Alstott, Steve Omohundro, Mark Waser, and two anonymous reviewers for comments on an earlier version of this paper, and to Stuart Armstrong and Anders Sandberg for helpful background discussion. Any remaining errors are the responsibility of the authors. Work on this paper was funded in part by a grant from the Future of Life Institute. Any opinions, findings or recommendations in this document are those of the authors and do not necessarily reflect views of the Global Catastrophic Risk Institute, Future of Life Institute, nor of others.